\def\year{2020}\relax
\documentclass[letterpaper]{article} 
\usepackage{aaai20}  
\usepackage{times}  
\usepackage{helvet} 
\usepackage{courier}  
\usepackage[hyphens]{url}  
\usepackage{graphicx} 
\usepackage{graphicx}  
\usepackage{amsmath}
\nocopyright
\usepackage{algorithm}
\usepackage{algorithmic}
\usepackage{booktabs}
\usepackage{amssymb}
\usepackage{pifont}
\usepackage{xcolor}
\newcommand{\cmark}{\ding{51}}%
\definecolor{darkblue}{HTML}{1F4E79}
\definecolor{lightblue}{HTML}{00B0F0}
\definecolor{salmon}{HTML}{FF9C6B}
\definecolor{xiaomi}{HTML}{FF670A}
\definecolor{ccblue}{HTML}{2196F3}
\definecolor{ccyellow}{HTML}{FFAC13}
\definecolor{ccpink}{HTML}{E53935}
\definecolor{ccgreen}{HTML}{83C44E}
\definecolor{ccgray}{HTML}{A9A9AB}
\definecolor{ccorange}{HTML}{C6EC75}

\usepackage{tikz,subfigure}
\usetikzlibrary{decorations.pathreplacing,positioning}
\usepackage{xcolor}
\usetikzlibrary{spy}
\usetikzlibrary{shapes.geometric, arrows, backgrounds, fit}
\tikzstyle{layer} = [rectangle, minimum width=1cm, minimum height=0.5cm,text centered, draw=black, fill=lightblue]
\tikzstyle{cell} = [rectangle, minimum width=1cm, minimum height=0.5cm,text centered, draw=black, fill=salmon]
\tikzstyle{darklayer} = [rectangle, minimum width=0.5cm, minimum height=0.5cm,text centered, draw=black, fill=darkblue]
\tikzstyle{state} = [circle, inner sep=0pt, minimum width=0.5cm, minimum height=0.5cm,text centered, draw=black, fill=darkblue]
\tikzstyle{arrow} = [thick,->,>=stealth]
\tikzstyle{op} = [circle, minimum width=0.5cm, minimum height=0.5cm,text centered, draw=black]
\tikzstyle{cell_r} = [rectangle, minimum width=1cm, minimum height=0.5cm,text centered, fill=ccyellow,rotate=90]
\tikzstyle{cell_r_p} = [cell_r, fill=ccpink] %
\tikzstyle{cell_r_b} = [cell_r, fill=ccblue]
\tikzstyle{cell_r_y} = [cell_r, fill=ccyellow]
\tikzstyle{cell_r_g} = [cell_r, fill=ccgreen]
\tikzstyle{cell_se} = [draw=darkblue]
\tikzstyle{e3} = [minimum width=2cm]
\tikzstyle{e6} = [minimum width=3cm]

\title{Searching Beyond MobileNetV3 }
\author{\Large \textbf{Xiangxiang Chu, Bo Zhang, Ruijun Xu}\\
\\
\Large Xiaomi AI Lab\\ 
{ \tt \{chuxiangxiang, zhangbo11, xuruijun\}@xiaomi.com}\\
}
 
\begin{document}
\maketitle

\begin{abstract}
The evolution of MobileNets has laid a solid foundation for neural network applications on mobile end. With the latest MobileNetV3, neural architecture search again claimed its supremacy in network design. Unfortunately, till today all mobile methods mainly focus on CPU latencies instead of GPU, the latter, however, is much preferred in practice for it has faster speed, lower overhead and less interference. Bearing the target hardware in mind, we propose the first Mobile GPU-Aware (MoGA) neural architecture search in order to be precisely tailored for real-world applications. Further, the ultimate objective to devise a mobile network lies in achieving better performance by maximizing the utilization of bounded resources. Urging higher capability while restraining time consumption is not reconcilable. We alleviate the tension by weighted evolution techniques. Moreover, we encourage increasing the number of parameters for higher representational power. With \textbf{200$\times$} fewer GPU days than MnasNet,  we obtain a series of models that outperform MobileNetV3 under the similar latency constraints, i.e., MoGA-A achieves \textbf{75.9\%} top-1 accuracy on ImageNet, MoGA-B meets 75.5\% which costs only 0.5 ms more on mobile GPU. MoGA-C best attests GPU-awareness by reaching 75.3\% and being slower on CPU but faster on GPU.  The models and test code are made publicly here \footnote{\url{https://github.com/xiaomi-automl/MoGA}}\footnote{This is a preview version, subject to frequent changes.}.
\end{abstract}

\section{Introduction}

The MobileNets trilogy has opened a gate to on-device artificial intelligence for the mobile vision world \cite{howard2017mobilenets,sandler2018mobilenetv2,howard2019searching}. In the meantime, neural architecture search becomes the new engine to empower the future architecture innovation \cite{zoph2018learning,tan2018mnasnet,cai2018proxylessnas,chu2019fairnas}. The guideline in designing mobile architecture is that not only should the high performance be concerned, but also we must strive for low latency in favor of rapid responsiveness and improved power efficiency to prolong battery life. 

\begin{figure}[ht]
\centering
{
\includegraphics[scale=0.6]{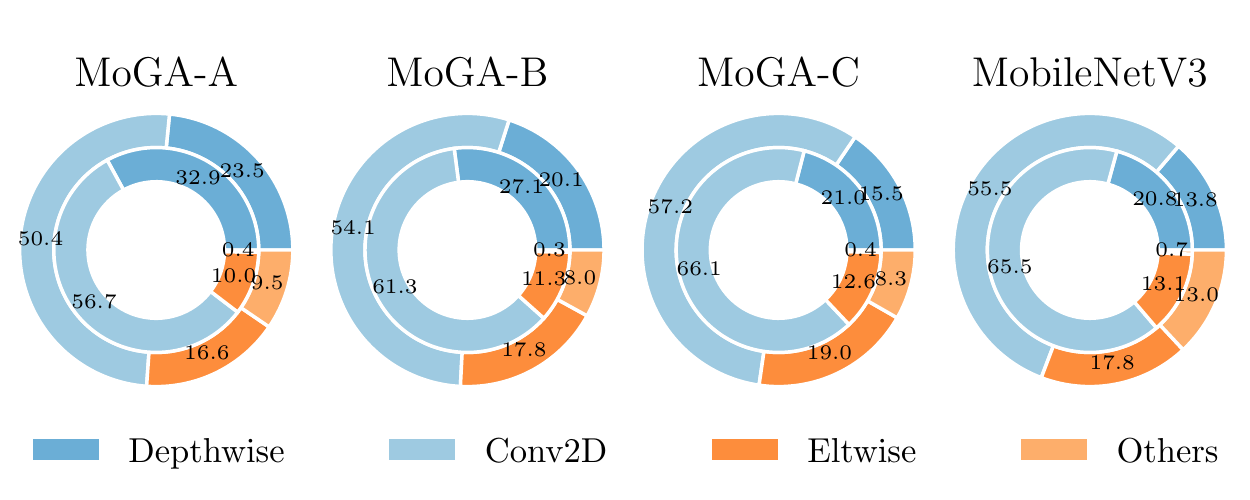}
}
\caption{Latency pie chart of MoGA-A/B/C, MobileNetV3 operations when run on mobile CPUs (inner circle with TFLite) vs. on mobile GPUs (outer circle with MACE).}
\label{fig:pie-a-ops}
\end{figure}

In this paper, we aim to bring forward the frontier of mobile neural architecture design by stretching out the representational space within the desired latency range. Our work can be summarized in following aspects.

First, we make a shift in the search trend from mobile CPUs to mobile GPUs, with which we can gauge the speed of a model more accurately and provide a production-ready solution. On this account, our overall search approach is named Mobile GPU-Aware neural architecture search (MoGA). Our results suggest that generated models show different behavior related to the targeted hardware as shown in Figure~\ref{fig:pie-a-ops}. 

Second, we replace traditional multi-objective optimization with a weighted fitness strategy. While considering accuracy, latency and the number of parameters as our objectives, particular care is required to abate these three contending forces. One important insight is that the number of parameters should be made reasonably large instead of as few as possible, this leverages performance but doesn't necessarily increase latency. At the mobile scale, this would be the proper choice as we try to avoid underfitting instead of overfitting. On top of that, we lay more attention on accuracy and latency than the number of parameters. 

Third, as per search cost, we benefit from one-shot supernet training and an accurate latency look-up table. Actually, it only requires the same expense as training a stand-alone model. The overall pipeline costs \textbf{12 GPU days}, about \textbf{200$\times$} less than MnasNet \cite{tan2018mnasnet}. More importantly, to cater for various mobile devices, as we decouple search process from training, it only requires one more inexpensive search with a renewed latency table.  In contrast, gradient descent and reinforced methods have to start all over for supernet training or incomplete training for multitudinous models \cite{liu2018darts,tan2018mnasnet}.

Finally, we present our searched architectures that outperform MobileNetV3. MoGA-A that achieves 75.9\% top-1 accuracy on ImageNet, MoGA-B 75.5\% and MoGA-C 75.3\%. MoGA-C is best comparable to MobileNetV3, with similar FLOPs and an equal number of parameters, which runs slower on mobile CPUs but faster on mobile GPUs.


\section{Related Works} 
\label{sec:related}

During the era of human craftsmanship, MobileNetV1 and V2 \cite{howard2017mobilenets,sandler2018mobilenetv2} have widely disseminated depthwise separable convolutions and inverted residuals with linear bottlenecks. Moreover, Squeeze and excitation blocks are later introduced in \cite{hu2018squeeze} to enrich residual modules from ResNet \cite{he2016deep}.

In their aftermath, a series of automated architectures are searched based on these building blocks \cite{tan2018mnasnet,cai2018proxylessnas,chu2019fairnas,howard2019searching}. For instance, MnasNet frames a factorized hierarchical search space with MobileNetV2's inverted bottleneck convolution blocks (MB) of variable kernel sizes and expansion rates. Its latest variation also includes an option of squeeze and excitation module (SE)  \cite{tan2018mnasnet}. ProxylessNAS and FairNAS adopt a similar design \cite{cai2018proxylessnas,chu2019fairnas} without SE modules, while MobileNetV3 achieves a new state of the art by integrating SE within MnasNet search space, along with numerous techniques like Platform-Aware NAS \cite{tan2018mnasnet},  NetAdapt \cite{yang2018netadapt} and improved non-linearities \cite{howard2019searching}.

As for search methods, recent attention has been drawn to the one-shot approaches initiated by \cite{bender2018understanding}, as they tremendously reduce computing resources and also offer state of the art results \cite{cai2018proxylessnas,stamoulis2019single,guo2019single,chu2019fairnas}. Briefly, a one-shot approach embodies weight-sharing across models by constructing a supernet where each step of training accounts for the final performance. Its single-path variations further cut down memory consumption by training a picked path at each step instead of the whole supernet, yielding more flexibility for architecture design \cite{stamoulis2019single,guo2019single,chu2019fairnas}. Among them, FairNAS proved it is critical to maintaining strict fairness for training single-path nets so to reach a steady rank, which can reasonably facilitate the search process \cite{chu2019fairnas}.

\section{Mobile GPU-Aware NAS Based on Multi-Objective Optimization}
\label{sec:gpu}

In this section, to better formulate our design problem, we draw insights from the development of MobileNets and experiments on the mobile GPU/CPU relationship, as well as reviewing previous optimization approaches.  
 
\subsection{Mobile GPU Awareness}
\begin{figure}[ht]
\centering
{
\includegraphics[scale=0.6]{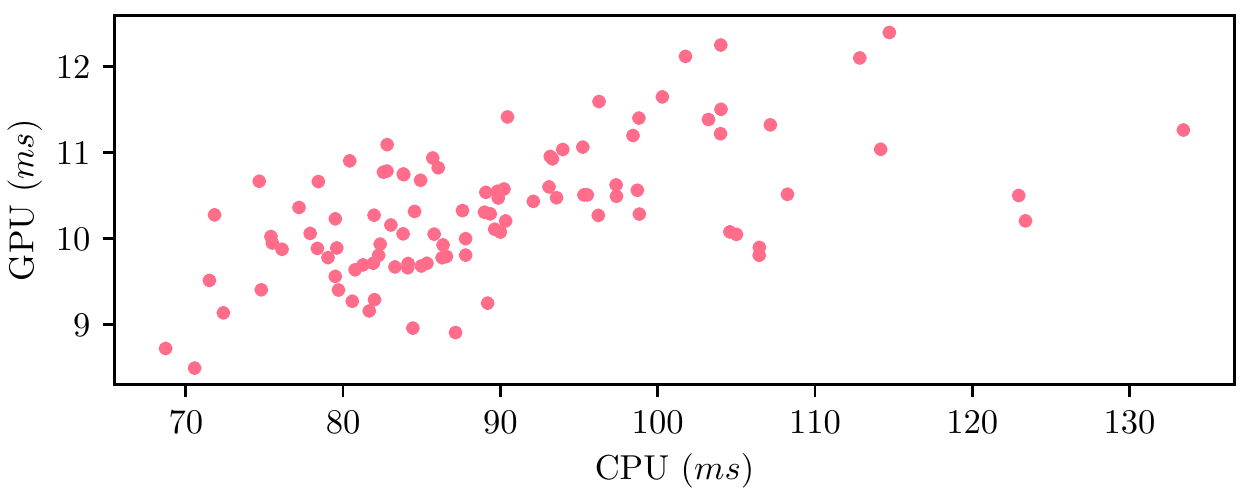}
}
\caption{Latency relationship on mobile CPUs vs. on mobile GPUs.}
\label{fig:latency_cpu_gpu}
\end{figure}
Recent NAS approaches have an increased emphasis on target platforms, primarily on mobile CPUs \cite{tan2018mnasnet,dong2018dpp,wu2018fbnet,cai2018proxylessnas,stamoulis2019single,howard2019searching}. MnasNet has developed a reward $ACC \times (LAT / TAR)^w$, which requires delicate manual tuning for a parameter $w$ to balance between  latency and accuracy \cite{tan2018mnasnet}, in MobileNetV3, $w$ is reduced from $-0.07$ to $-0.15$ \cite{howard2019searching} to compensate for accuracy drop.

In practice, mobile neural networks are mostly deployed to run on GPUs, DSPs and recently also on specific Neural Processing Units (NPUs), while CPUs would be the last to choose. 
To further investigate the relationship of CPU latencies versus GPU ones, we measure 100 random models on both two platforms. The result is shown in Figure~\ref{fig:latency_cpu_gpu}. We see that there is no obvious linear correspondence. Hence, 
To develop architectures with target hardware in mind is more than necessary. For this reason, we are driven to apply Mobile GPU awareness to the latest neural architecture search approaches. 


\subsection{Underfitting and Overfitting}
\label{sec:more_params}
\begin{figure}[ht]
\centering
{
\includegraphics[scale=0.6]{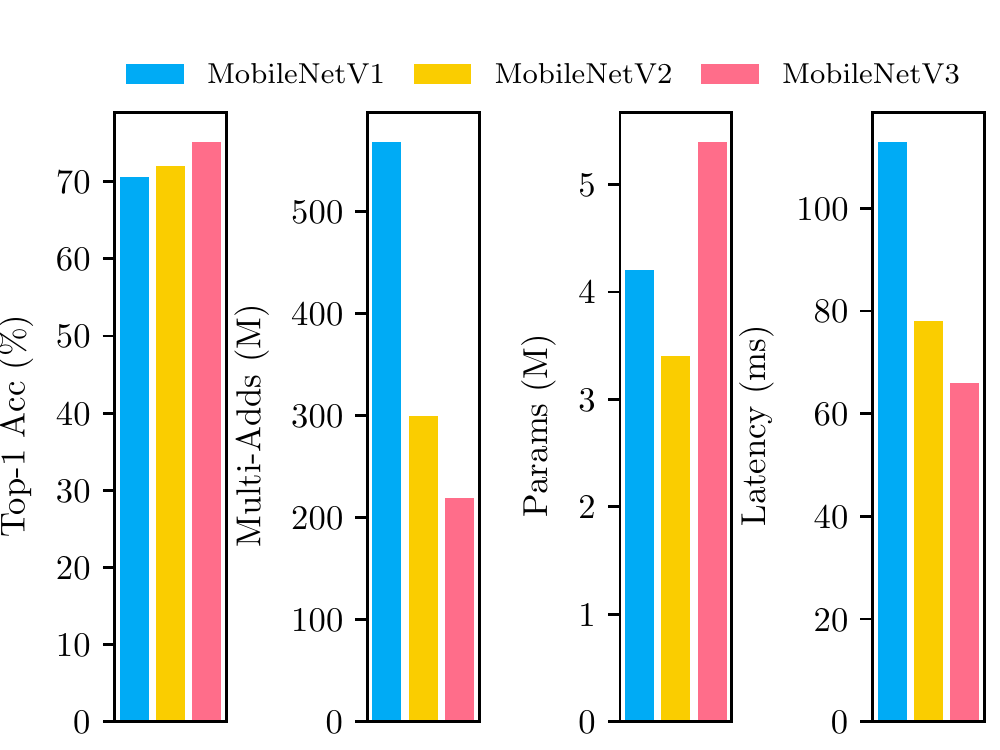}
}
\caption{The evolution of MobileNets.}
\label{fig:mobilenet-comp}
\end{figure}

As we try to tear apart two contradicting objectives, there isn't too much freedom left to increase accuracy with a constrained latency. Fortunately, we observe from the evolution of MobileNets as in Figure~\ref{fig:mobilenet-comp}, the number of parameters has grown while the latencies and multiply-adds are kept low.

Moreover, for the mobile end, models tend to be underfitted instead of overfitted since they carry fewer numbers of parameters \cite{zhang2018shufflenet}, which means we are free to encourage representational power by enlarging its range of parameters. This intuition greatly expands our design space.

\subsection{Problem Formulation}
\label{sec:prob}
Most hardware-aware methods build the classification problem as follows,
\begin{equation}
\label{eq: p1}
\begin{split}
\max & \quad   Accuracy(m) \\
s.t. & \quad Latency(m) < L \\
& \quad \text{model} \, m \in \Omega.
\end{split}
\end{equation}
where $\Omega$ is the whole search space and $L$ is a given maximal acceptable latency\footnote{The following latency means mobile GPU latency.}. Informally speaking, larger models have greater capacity and to achieve better accuracy. Therefore, NAS methods will prefer models which have large running time.  As a result, when the requirement of $L$ changes, the whole NAS pipeline must start over.

To address the above problem, a recent popular approach formulate it as multi-objective problem (MOP) whose solution is called Pareto Front,

\begin{equation}
\label{eq: two-obj}
\begin{split}
max & \quad \{Accuracy(m), -Latency(m)\} \\
& \quad m \in \Omega.
\end{split}
\end{equation}

One popular approach for Equation~\ref{eq: two-obj} is converting it into a customized objective of weighted product  $ACC \times (LAT / TAR)^w$, which requires delicate manual tuning for a parameter $w$ to balance between  latency and accuracy \cite{tan2018mnasnet}.

Upon these previous attempts, as inspired by Section~\ref{sec:more_params}, we maximize the number of parameters in addition to the two objectives in Equation~\ref{eq: two-obj}. More formally, we try to solve the following problem, 

\begin{equation}
	\begin{split}
	max & \quad \{ Accuracy(m), -Latency(m),Params(m)\} \\
	& \quad m \in \Omega.
	\end{split}
\end{equation}

As a matter of fact, these three objectives are not of equal importance in most cases. A typical case is like Equation~\ref{eq: two-obj}. Therefore, we need to introduce some strategies to address the issue. Let $w_{acc}, w_{lat}, w_{params}$ denote customized preference for those objectives. Without loss of generality, the problem can be defined as,

\begin{equation}
\label{eq:three_objs}
\begin{split}
min & \quad \{ -Accuracy(m), Latency(m), -Params(m)\} \\
s.t. & \quad m \in \Omega \\
&\quad w_{acc} + w_{lat}  + w_{params} = 1 \\
&\quad  w_{acc},w_{lat}, w_{params}  >=0.
\end{split}
\end{equation}

There are two basic subproblems to be solved in the next section. One is to instantly evaluate accuracy and latency of a model, the other is  to solve Equation~\ref{eq:three_objs}. We use NSGA-II, which is one of the most powerful and widely used algorithms to solve such problems \cite{deb2002fast}. First, it's efficient to solve MOPs, especially when the number of objectives is large, some variants can still work. Second, it's flexible to apply customized preferences for different objectives, as well as various constraints. We also benefit from its implicit objective scaling and normalization.

\section{Solving it Using Weighted NSGA-II}
\label{sec:solve}

\subsection{Search Space}
Our search space is built layer by layer on inverted bottleneck blocks as \cite{cai2018proxylessnas,chu2019fairnas}. We keep the same number of layers and activation functions as MobilenetV3-large. For each layer, we search from three dimensions (see Table~\ref{tab:layer-choice}):
\begin{itemize}
	\item the convolution kernel size (3, 5, 7)
	\item the expansion ratio for the inverted bottleneck block (3, 6)
	\item whether the squeezing and excitation mechanism is enabled or not.
\end{itemize}

Therefore, the total search space has a volume of $12^{14}$, which needs efficient methods to differentiate better models from worse.
To be simple, we search for the expansion rate instead of channels which is  used by \cite{howard2019searching} based on NetAdapt \cite{yang2018netadapt}. Besides, we utilize choice index to directly encode each model chromosome. More formally, a model chromosome  $m$ can be written as $m_1 = (x^1_1,x^1_2,...,x^1_{14})$.

\begin{table}[]
\centering{
\begin{small}
\begin{tabular}{*{4}{l}}
\toprule
Index & Expansion & Kernel Size & SE \\
\midrule
0 & 3 & 3 & -\\
1 & 3 & 3 & \cmark \\
2 & 3 & 5 & -\\
3 & 3 & 5 & \cmark \\
4 & 3 & 7 & -\\
5 & 3 & 7 & \cmark \\
6 & 6 & 3 & -\\
7 & 6 & 3 & \cmark \\
8 & 6 & 5 & -\\
9 & 6 & 5 & \cmark \\
10 & 6 & 7 & -\\
11 & 6 & 7 & \cmark \\
\bottomrule
\end{tabular}
\end{small}
}
\caption{Each layer in our search space has 12 choices. SE: Squeeze-and-Excitation.}
\label{tab:layer-choice}
\end{table}

\subsection{Accuracy and Latency Prediction}
\label{sec:acc}
The evaluation of model accuracy must be made immediate for searching efficiency. We take advantage of a variation of one-shot approaches FairNAS for fast evaluation with a stable ranking. Unlike their version, based on our previously defined search space, we construct a supernet with 12 choice blocks per layer. Then we train our supernet on the ImageNet dataset with the same fairness strategy. 

As for mobile GPU latency, we don't acquire real-time latency during the pipeline from a cell phone for two reasons. One is that while the performance can be rapidly predicted by the supernet which takes less than 1 minute, it will easily become the bottleneck when we use a mobile device to evaluate latency on the fly. The other is that latency measurement may become inaccurate as a result of overheating after long-time insistent testing.

\begin{figure}[ht]
\centering
{
\includegraphics[scale=0.6]{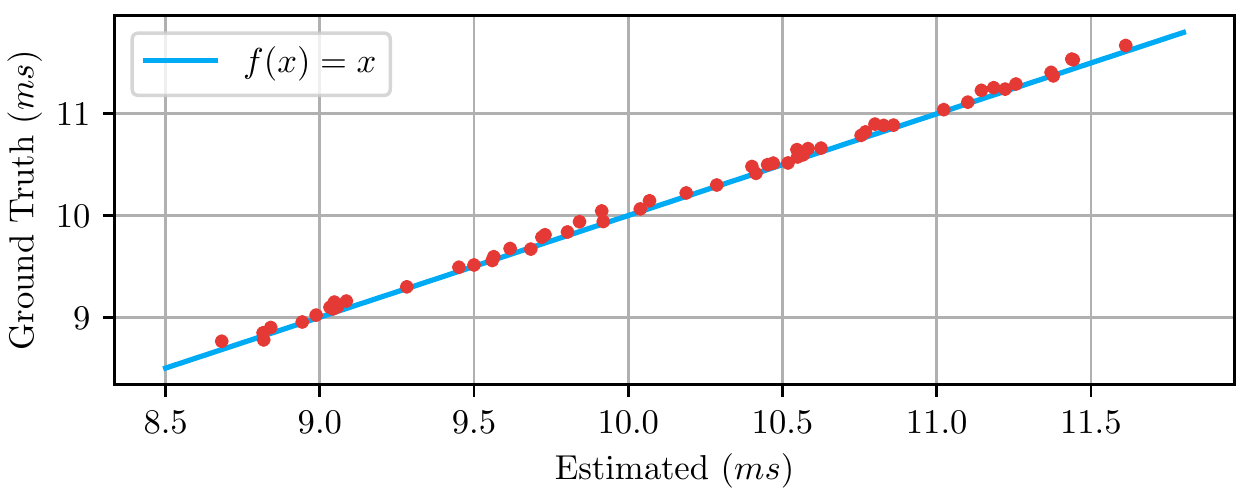}
}
\caption{Mobile GPU latency measured vs. predicted ones. The latency RMSE is 0.0571ms. }
\label{fig:mobile-latency}
\end{figure}

Instead, since each choice block in our search space has a fixed input, we can efficiently approximate the latency for any sampled model. To do so, we benchmark the latency of each choice block under a given input and construct a layerwise lookup table. We can then accurately calculate the latency simply by accumulating time cost across all layers. We find that predicted GPU latency coincides with ground-truth values with a negligible RMSE, see Figure~\ref{fig:mobile-latency}.

\subsection{Weighted NSGA-II}
We comply with the standard NSGA-II procedure, and only state the difference if necessary.
\subsubsection{Population Initialization}
We initialize population to introduce various choice blocks to encourage exploration.
\subsubsection{Crossover}
We take a single-point crossover. Specifically, for two models  $m_1 = (x^1_1,x^1_2,...,x^1_{14})$ and $m_2 = (x^2_1,x^2_2,...,x^2_{14})$, if the point position is $k$, the result after crossover is
 $(x^1_1,x^1_2,...x^2_k...,x^1_{14})$.
\subsubsection{Mutation}
We use hierarchical mutation and the same hyperparameters as FairNAS \cite{chu2019fairnas}.
\subsubsection{Non-dominated Sorting}
For a minimization problem with $n$ objectives, we state that $A$ dominates $B$ means for any objective $O^i$, $O_A^i \le O_B^i$. For a given population $P$, $A$ is not dominated if and only if $A$ is not dominated by any other individuals.

The crowding distance is a key component to achieve better trade-off among various objectives. We use the customized weights to define the crowding distance for non-boundary individuals,
\begin{equation}
\label{eq: distance}
	\begin{split}
	D(m_j) = \sum_{i=1}^{n}w_i*\frac{O^i_{neighbor+}-O^i_{neighbor-}}{O^i_{max}-O^i_{min}}.
	\end{split}
\end{equation}

Note $O^i_{neighbour-}$ and $O^i_{neighbor+}$  are the $i$-th objective value of the left and the right neighbor of model $m_j$ respectively, while $O^i_{max}$ and $O^i_{min}$ are the maximum and the minimum for the $i$-th objective in the current population.
In Equation~\ref{eq: distance}, customized preference can be flexibly incorporated. If $w_{acc} = w_{lat}  =  w_{params} =\frac{1}{3}$, it degrades as the standard NSGA-II. In our experiment, $w_{acc} = w_{lat}=0.4, w_{params}=0.2$, whose requirement is from a practical application. 

\subsection{Our NAS Pipeline} 
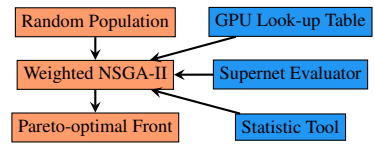
\begin{figure}[ht]
\centering
{
\begin{tikzpicture}[scale=0.7, every node/.style={scale=0.7}]
\node (a) [cell] {Random Population};
\node (b) [cell,below of=a] {Weighted NSGA-II};
\node (c) [cell,above right of=b, xshift=3cm, yshift=0.3cm,fill=ccblue] {GPU Look-up Table};
\node (d) [cell,below of=c,fill=ccblue] {Supernet Evaluator};
\node (e) [cell,below of=d,fill=ccblue] {Statistic Tool};
\node (f) [cell,below of=b] {Pareto-optimal Front};
\draw [arrow] (a) -- (b);
\draw [arrow] (b) -- (f);
\draw [arrow] (c) -- (b);
\draw [arrow] (d) -- (b);
\draw [arrow] (e) -- (b);
\end{tikzpicture}
}
\caption{The overall pipeline of MoGA.}
\label{fig:pipeline}
\end{figure}

Our search pipeline is an evolution process, drawn in Figure~\ref{fig:pipeline} and detailed in  Algorithm~\ref{alg:nas_pipeline}. Specifically, we use a trained supernet as a fast evaluator, a GPU latency lookup table, and a statistic tool to compute the number of parameters. The initial random population propagates at significant speed. The pipeline evolves 120 generations with a population size 70, and it only takes about 1.5 GPU days to evaluate these 8400 models. We use the same hyperparameters as FairNAS.

\begin{algorithm}[tb]
	\caption{The weighted NAS pipeline.}
	\label{alg:nas_pipeline}
	\begin{algorithmic}
		\STATE {\bfseries Input:} Supernet $S$, search space $\Omega$, the number of generations $N$, population size $n$, validation dataset $D$, objective weights $w$
		\STATE {\bfseries Output: } A set of $K$ individuals on the Pareto front.
		\STATE {Train supernet $S$ by the FairNAS approach on $\Omega$.}
		\STATE {Make gpu latency table \emph{T} as section \ref{sec:acc}}.
		
		\STATE {Uniformly generate the populations $P_0$ and $Q_0$ until each has $n$ individuals.}
		\FOR {$i=0$ {\bfseries to} $N-1$}
		\STATE{$R_i = P_i \cup Q_i$}
		\STATE {$F = \text{non-dominated-sorting}(R_i)$}
		\STATE{Pick $n$ individuals to form $P_{i+1}$ by ranks and the crowding distance \textbf{weighted} by $w$ based on Equation~\ref{eq: distance}.}
		\STATE {$Q_{i+1} = \emptyset$}
		\WHILE {$size(Q_{i+1}) < n $}

		\STATE {$ M = \text{tournament-selection}(P_{i+1})$}

		\STATE {$q_{i+1} = \text{crossover}(M) \cup \text{hierarchical-mutation}(M)  $ }

        \STATE {Obtain fitness value across all objectives}
		 \STATE {\quad  Evaluate model $q_{i+1}$'s accuracy with $S$ on $D$} 
		\STATE { \quad Regress model $q_{i+1}$'s latency based on  \emph{T}} 

		\STATE {$Q_{i+1} = Q_{i+1} \cup \{q_{i+1}\}$ }
		\ENDWHILE
		
		\ENDFOR
		\STATE {Select $K$ models at an equal distance near Pareto front from $P_{N}$}
	\end{algorithmic}
\end{algorithm}

\begin{table}[]
\centering{
\begin{small}
\begin{tabular}{*{7}{l}}
\toprule
Input & Ops & $t$ & $c$ & SE & NL & $s$  \\
\midrule
$224^2 \times 3$ &conv2d, $3\times3$ & - & 16  &- & HS &2  \\
$112^2 \times 16$ & bneck, $3\times3$  & 1  &16  &- & RE  &1  \\
$112^2 \times 16$ & bneck, $5\times5$ & 6  &24  &- & RE &2  \\
$56^2 \times 24$ & bneck, $7\times7$ & 6 &24 &- & RE &1 \\
$56^2 \times 24$ & bneck, $3\times3$ & 6 &40 & -& RE &2 \\
$28^2 \times 40$ & bneck, $3\times3$ &  6  &40 & \cmark & RE&1 \\
$28^2 \times 40$ & bneck, $3\times3$ & 3  &40 &\cmark & RE&1 \\
$28^2 \times 40$ & bneck, $3\times3$ & 6  &80 & \cmark & HS &2 \\
$14^2 \times 80$ & bneck, $3\times3$ & 6  &80 & - & HS &1 \\
$14^2 \times 80$ & bneck, $7\times7$ & 6  &80 & - & HS &1 \\
$14^2 \times 80$ & bneck, $7\times7$ & 3  &80 &  \cmark & HS &1 \\
$14^2 \times 80$ & bneck, $7\times7$ & 6  &112 & - & HS &1 \\
$14^2 \times 112$ & bneck, $3\times3$ & 6  &112 & - & HS &1 \\
$14^2 \times 112$ & bneck, $3\times3$ & 6  &160 & - & HS &2 \\
$7^2 \times 160$ & bneck, $5\times5$ & 6  &160 &\cmark & HS & 1 \\
$7^2 \times 160$ & bneck, $5\times5$ & 6  &160 &\cmark & HS &1 \\
$7^2 \times 160$ & conv2d, $1\times1$ & -  &960 &- & HS& 1 \\
$7^2 \times 960$ & avgpool, $7\times7$ & -  &- &- & HS &- \\
$1^2 \times 960$ & conv2d, $1\times1$ & -  & 1280 &- & HS &1\\
$1^2 \times 1280$ & conv2d, $1\times 1$ & -  &k & - &- & -\\
\bottomrule
\end{tabular}
\end{small}
}
\caption{The architecture of MoGA-A. Note $t,c,s$ refer to expansion rate, output channel size and stride respectively. SE for squeeze-and-excitation, NL for non-linearity. $k$ for the number of categories.}
\label{tab:moga-a}
\end{table}

\begin{table*}
	\begin{center}
	  \begin{small}
		\begin{tabular}{*{8}{l}} 			
		\toprule
			Methods & Mult-Adds  & Params & Lat$_g^{SNPE}$&  Lat$_{g}^{MACE}$& Lat$_{c}$ & Top-1  & Top-5  \\
			& (M) & (M) & (ms) & (ms) & (ms) & (\%) & (\%) \\
			\midrule
			MobileNetV2 1.0 \cite{sandler2018mobilenetv2}& 300 & 3.4 & 6.9$^\dagger$ & 7.0$^\dagger$ & 78 & 72.0 & 91.0\\ 
			MobileNetV3 Large 1.0  \cite{howard2019searching} & 219 & 5.4 & 10.8$^\star$ & 9.5$^\star$ &70 (66)$^\star$ &75.0 (75.2)$^\star$ & 92.2 \\ 
			\midrule
			MnasNet -A1 \cite{tan2018mnasnet} & 312 & 3.9 &- &- & 78 & 75.2 & 92.5 \\ 
			MnasNet-A2 \cite{tan2018mnasnet}  & 340 & 4.8 &- & -& 84 & 75.6 & 92.7 \\ 
			FBNet-B \cite{wu2018fbnet} & 295 & 4.5 &- & -& 23 $^\ddagger$ & 74.1 & - \\ 
			Proxyless-R Mobile \cite{cai2018proxylessnas}   & 320$^\dagger$ & 4.0 & 7.3$^\dagger$ & 7.9$^\dagger$ & 87 (78)$^\dagger$ & 74.6 & 92.2 \\  
			Proxyless GPU \cite{cai2018proxylessnas}  & 465$^\dagger$ & 7.1 & 9.6$^\dagger$ & 9.8$^\dagger$ & 126 (124)$^\dagger$ & 75.1 & -\\ 
			Single-Path NAS \cite{stamoulis2019single} & 365 & 4.3 & -& -& 79 & 75.0 & 92.2 \\ 
			Once for All \cite{cai2019once} & 327& - & - & - & 112 $^*$ & 75.3 & - \\
			 FairNAS-A \cite{chu2019fairnas}  & 388& 4.6 & 9.8$^\dagger$ &9.7$^\dagger$ & 104 & 75.3 & 92.4 \\ 
			 MoGA-A (Ours) & 304 & 5.1 & 11.8 & 11.1 &  101 & \textbf{75.9} & 92.8 \\
			 MoGA-B (Ours) & 248 & 5.5 & 10.3 & 10.0 &  81 & 75.5 & 92.6 \\
			 MoGA-C (Ours) & 221 & 5.4 & 9.6 & 8.8 &   71 & 75.3  & 92.5 \\
			\bottomrule
		\end{tabular}
		\end{small}
	\end{center}
	\caption{Comparison of mobile models on ImageNet.  $^\star$: Our reimplementation. Numbers within the parentheses are reported by its authors, same for below. $^\dagger$: Based on its published code. $^\ddagger$: Samsung Galaxy S8. $^*$: Samsung Note8.} 
	\label{tab:comparison-imagenet}
\end{table*}

\section{Experiments}
\label{sec:exp}

\subsection{Mobile GPU Latency}
In practice, we employ SNPE \cite{qualcomm2019snpe}  and Mobile AI Compute Engine (MACE) for mobile GPU benchmarking \cite{xiaomi2018mace}. We randomly sample some models  and report the differences between our predictions and on-device measurements, which are shown in Figure~\ref{fig:mobile-latency}. For instant latency prediction, we construct a latency lookup table based on MACE measurements on all 12 choices blocks for each cell ($12 \times 14$). For the final comparison with state-of-the-art models, we also report mobile GPU latencies with SNPE \cite{qualcomm2019snpe}, and CPU latencies with Tensorflow Lite \cite{tensorflow2015-whitepaper}. Considering recent updates on Tensorflow speed up the inference time, we choose a version that can reproduce the result on MobileNetV2 \cite{sandler2018mobilenetv2}. 

Unless otherwise noted, mobile CPU latencies are measured on a Google Pixel 1 using a single large core of CPU with a batch size of 1. Mobile GPU latencies are benchmarked on a Mi MIX 3. The input size is set to 224$\times$224.

\subsection{Training}
\subsubsection{Training of our Supernet}
We search proxylessly on the ImageNet \cite{deng2009imagenet} classification dataset. We take out 50k images  from the training set to form our validation set and use the official validation set as our test set to evaluate our models, which is on par with other methods. In particular, we train the supernet by SGD with momentum 0.9 for 32 epochs.  The initial learning rate is 0.05 and is scheduled to arrive at zero within a single cosine cycle. 


\subsubsection{Training for Stand-Alone models}
To alleviate the training unfairness, we utilize the same training tricks and hyperparameters as MobileNetV3 \cite{howard2019searching}.6. By doing so, we singled out various training tricks in order to focus on the authentic model performance. In particular, We use a batch size of 4096 and RMSProp optimizer with 0.9 momentum. The initial learning rate is 0.1 and linear warm-up \cite{goyal2017accurate} is applied for the first 5 epochs. We use a dropout rate of 0.2 before the last layer \cite{srivastava2014dropout} and L2 weight decay $1e-5$. Besides, we make use of NVIDIA's mixed precision library Apex to enable larger batch size\footnote{https://github.com/NVIDIA/apex.git}. All our experiments are performed on two Tesla-V100 machines.

\subsection{Comparisons with State-of-the-art Methods}
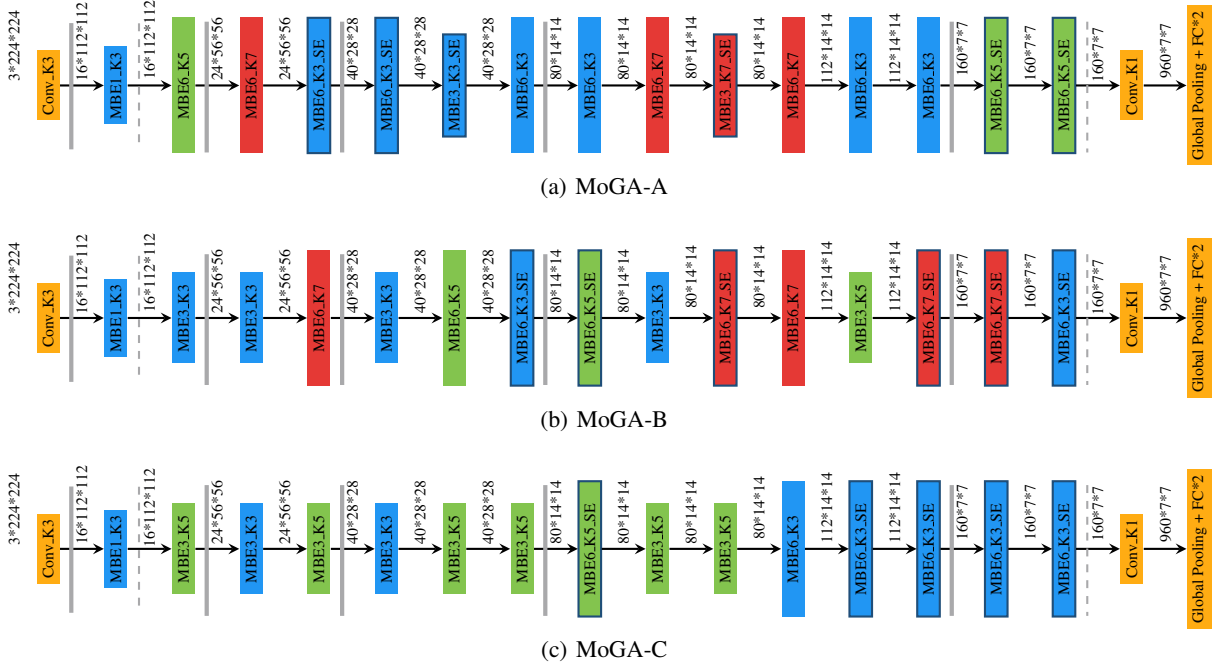
\begin{figure*}[ht]
\begin{center}
\subfigure[MoGA-A]{
\centerline{
    \begin{tikzpicture}[thick,scale=0.6, every node/.style={scale=0.6},node distance=1.5cm]
    	\node (s1) [cell_r,label={[rotate=90,xshift=0.1cm,yshift=0.5cm]right:3*224*224}] {Conv\_K3};
	\node (s2) [cell_r_b, yshift=-1.5cm, xshift=-1.5cm, right of=s1,label={[rotate=90,xshift=0.1cm,yshift=0.5cm]right:16*112*112}] {MBE1\_K3};
        \node (a) [cell_r_g, e6, yshift=-1.5cm, xshift=-1.5cm, right of=s2,label={[rotate=90,xshift=-0.5cm,yshift=0.5cm]right:16*112*112}] {MBE6\_K5}; 
        \node (b) [cell_r_p,e6 , yshift=-1.5cm, xshift=-1.5cm, right of=a,label={[rotate=90,xshift=-0.7cm,yshift=0.5cm]right:24*56*56}] {MBE6\_K7};
        \node (c) [cell_r_b,e6, cell_se , yshift=-1.5cm, xshift=-1.5cm, right of=b,label={[rotate=90,xshift=-0.7cm,yshift=0.5cm]right:24*56*56}] {MBE6\_K3\_SE}; 
        \node (d) [cell_r_b,e6, cell_se, yshift=-1.5cm, xshift=-1.5cm, right of=c,label={[rotate=90,xshift=-0.7cm,yshift=0.5cm]right:40*28*28}] {MBE6\_K3\_SE};
        \node (e) [cell_r_b,e3, cell_se , yshift=-1.5cm, xshift=-1.5cm, right of=d,label={[rotate=90,xshift=-0.3cm,yshift=0.5cm]right:40*28*28}] {MBE3\_K3\_SE};
        \node (f) [cell_r_b, e6 , yshift=-1.5cm, xshift=-1.5cm, right of=e,label={[rotate=90,xshift=-0.7cm,yshift=0.5cm]right:40*28*28}] {MBE6\_K3};
        \node (g) [cell_r_b, e6 , yshift=-1.5cm, xshift=-1.5cm, right of=f,label={[rotate=90,xshift=-0.7cm,yshift=0.5cm]right:80*14*14}] {MBE6\_K3};
        \node (h) [cell_r_p, e6 , yshift=-1.5cm, xshift=-1.5cm, right of=g,label={[rotate=90,xshift=-0.7cm,yshift=0.5cm]right:80*14*14}] {MBE6\_K7};
        \node (i) [cell_r_p, e3 , cell_se, yshift=-1.5cm, xshift=-1.5cm, right of=h,label={[rotate=90,xshift=-0.3cm,yshift=0.5cm]right:80*14*14}] {MBE3\_K7\_SE};
        \node (j) [cell_r_p, e6 , yshift=-1.5cm, xshift=-1.5cm, right of=i,label={[rotate=90,xshift=-0.7cm,yshift=0.5cm]right:80*14*14}] {MBE6\_K7};
        \node (k) [cell_r_b, e6 , yshift=-1.5cm, xshift=-1.5cm, right of=j,label={[rotate=90,xshift=-0.7cm,yshift=0.5cm]right:112*14*14}] {MBE6\_K3};
        \node (l) [cell_r_b, e6 , yshift=-1.5cm, xshift=-1.5cm, right of=k,label={[rotate=90,xshift=-0.7cm,yshift=0.5cm]right:112*14*14}] {MBE6\_K3};
        \node (m) [cell_r_g, e6 , cell_se, yshift=-1.5cm, xshift=-1.5cm, right of=l,label={[rotate=90,xshift=-0.7cm,yshift=0.5cm]right:160*7*7}] {MBE6\_K5\_SE}; 
        \node (n) [cell_r_g, e6 , cell_se, yshift=-1.5cm, xshift=-1.5cm, right of=m,label={[rotate=90,xshift=-0.8cm,yshift=0.5cm]right:160*7*7}] {MBE6\_K5\_SE};
        \node (bp) [cell_r,yshift=-1.5cm, xshift=-1.5cm, right of=n,label={[rotate=90,xshift=-0.1cm,yshift=0.5cm]right:160*7*7}] {Conv\_K1};
        \node (pool_fc) [cell_r,yshift=-1.5cm, xshift=-1.5cm, right of=bp,label={[rotate=90,xshift=-1cm,yshift=0.5cm]right:960*7*7}] {Global Pooling + FC*2};
        \draw [arrow] (s1) -- (s2);
        \draw [arrow] (s2) -- (a);
        \draw [arrow] (a) -- (b);
        \draw [arrow] (b) -- (c);
        \draw [arrow] (c) -- (d);
        \draw [arrow] (d) -- (e);
        \draw [arrow] (e) -- (f);
        \draw [arrow] (f) -- (g);
        \draw [arrow] (g) -- (h);
        \draw [arrow] (h) -- (i);
        \draw [arrow] (i) -- (j);
        \draw [arrow] (j) -- (k);
        \draw [arrow] (k) -- (l);
        \draw [arrow] (l) -- (m);
        \draw [arrow] (m) -- (n);
        \draw [arrow] (n) -- (bp);
        \draw [arrow] (bp) -- (pool_fc);
        \draw [ultra thick, ccgray] (s1.south east)+(0.25,0.6) -- +(0.25,-2.2);
        \draw [ultra thick, ccgray] (a.south east)+(0.25,-0.1) --+(0.25,-3);
        \draw [ultra thick, ccgray] (c.south east)+(0.25,-0.1) --+(0.25,-3);
        \draw [ultra thick, ccgray] (f.south east)+(0.25,-0.1) --+(0.25,-3);
        \draw [ultra thick, ccgray] (l.south east)+(0.25,-0.1) --+(0.25,-3);
        \draw [dashed, ccgray] (s2.south east)+(0.25,0.5) -- +(0.25,-2.3);
       \draw [dashed, ccgray] (n.south east)+(0.25,-0.1) -- +(0.25,-3);  
    \end{tikzpicture}
}}
\subfigure[MoGA-B]{
\centerline{
    \begin{tikzpicture}[thick,scale=0.6, every node/.style={scale=0.6},node distance=1.5cm]
    	\node (s1) [cell_r,label={[rotate=90,xshift=0.1cm,yshift=0.5cm]right:3*224*224}] {Conv\_K3};
	\node (s2) [cell_r_b, yshift=-1.5cm, xshift=-1.5cm, right of=s1,label={[rotate=90,xshift=0.1cm,yshift=0.5cm]right:16*112*112}] {MBE1\_K3};
        \node (a) [cell_r_b, e3, yshift=-1.5cm, xshift=-1.5cm, right of=s2,label={[rotate=90,xshift=0cm,yshift=0.5cm]right:16*112*112}] {MBE3\_K3}; 
        \node (b) [cell_r_b,e3 , yshift=-1.5cm, xshift=-1.5cm, right of=a,label={[rotate=90,xshift=-0.2cm,yshift=0.5cm]right:24*56*56}] {MBE3\_K3};
        \node (c) [cell_r_p,e6 , yshift=-1.5cm, xshift=-1.5cm, right of=b,label={[rotate=90,xshift=-0.7cm,yshift=0.5cm]right:24*56*56}] {MBE6\_K7}; 
        \node (d) [cell_r_b,e3 , yshift=-1.5cm, xshift=-1.5cm, right of=c,label={[rotate=90,xshift=-0.2cm,yshift=0.5cm]right:40*28*28}] {MBE3\_K3};
        \node (e) [cell_r_g,e6 , yshift=-1.5cm, xshift=-1.5cm, right of=d,label={[rotate=90,xshift=-0.7cm,yshift=0.5cm]right:40*28*28}] {MBE6\_K5};
        \node (f) [cell_r_b, e6, cell_se , yshift=-1.5cm, xshift=-1.5cm, right of=e,label={[rotate=90,xshift=-0.7cm,yshift=0.5cm]right:40*28*28}] {MBE6\_K3\_SE};
        \node (g) [cell_r_g, e6, cell_se , yshift=-1.5cm, xshift=-1.5cm, right of=f,label={[rotate=90,xshift=-0.7cm,yshift=0.5cm]right:80*14*14}] {MBE6\_K5\_SE};
        \node (h) [cell_r_b, e3 , yshift=-1.5cm, xshift=-1.5cm, right of=g,label={[rotate=90,xshift=-0.2cm,yshift=0.5cm]right:80*14*14}] {MBE3\_K3};
        \node (i) [cell_r_p, e6 , cell_se, yshift=-1.5cm, xshift=-1.5cm, right of=h,label={[rotate=90,xshift=-0.6cm,yshift=0.5cm]right:80*14*14}] {MBE6\_K7\_SE};
        \node (j) [cell_r_p, e6, yshift=-1.5cm, xshift=-1.5cm, right of=i,label={[rotate=90,xshift=-0.6cm,yshift=0.5cm]right:80*14*14}] {MBE6\_K7};
        \node (k) [cell_r_g, e3 , yshift=-1.5cm, xshift=-1.5cm, right of=j,label={[rotate=90,xshift=-0.1cm,yshift=0.5cm]right:112*14*14}] {MBE3\_K5};
        \node (l) [cell_r_p, e6 , cell_se, yshift=-1.5cm, xshift=-1.5cm, right of=k,label={[rotate=90,xshift=-0.6cm,yshift=0.5cm]right:112*14*14}] {MBE6\_K7\_SE};
        \node (m) [cell_r_p, e6 , cell_se, yshift=-1.5cm, xshift=-1.5cm, right of=l,label={[rotate=90,xshift=-0.7cm,yshift=0.5cm]right:160*7*7}] {MBE6\_K7\_SE}; 
        \node (n) [cell_r_b, e6 , cell_se, yshift=-1.5cm, xshift=-1.5cm, right of=m,label={[rotate=90,xshift=-0.7cm,yshift=0.5cm]right:160*7*7}] {MBE6\_K3\_SE};
        \node (bp) [cell_r,yshift=-1.5cm, xshift=-1.5cm, right of=n,label={[rotate=90,xshift=-0.1cm,yshift=0.5cm]right:160*7*7}] {Conv\_K1};
        \node (pool_fc) [cell_r,yshift=-1.5cm, xshift=-1.5cm, right of=bp,label={[rotate=90,xshift=-1cm,yshift=0.5cm]right:960*7*7}] {Global Pooling + FC*2};
        \draw [arrow] (s1) -- (s2);
        \draw [arrow] (s2) -- (a);
        \draw [arrow] (a) -- (b);
        \draw [arrow] (b) -- (c);
        \draw [arrow] (c) -- (d);
        \draw [arrow] (d) -- (e);
        \draw [arrow] (e) -- (f);
        \draw [arrow] (f) -- (g);
        \draw [arrow] (g) -- (h);
        \draw [arrow] (h) -- (i);
        \draw [arrow] (i) -- (j);
        \draw [arrow] (j) -- (k);
        \draw [arrow] (k) -- (l);
        \draw [arrow] (l) -- (m);
        \draw [arrow] (m) -- (n);
        \draw [arrow] (n) -- (bp);
        \draw [arrow] (bp) -- (pool_fc);
        \draw [ultra thick, ccgray] (s1.south east)+(0.25,0.6) -- +(0.25,-2.2);
        \draw [ultra thick, ccgray] (a.south east)+(0.25,0.4) --+(0.25,-2.5);
        \draw [ultra thick, ccgray] (c.south east)+(0.25,-0.1) --+(0.25,-3);
        \draw [ultra thick, ccgray] (f.south east)+(0.25,-0.1) --+(0.25,-3);
        \draw [ultra thick, ccgray] (l.south east)+(0.25,-0.1) --+(0.25,-3);
        \draw [dashed, ccgray] (s2.south east)+(0.25,0.5) -- +(0.25,-2.3);
       \draw [dashed, ccgray] (n.south east)+(0.25,-0.1) -- +(0.25,-3);  
    \end{tikzpicture}
}}
\subfigure[MoGA-C]{
\centerline{
    \begin{tikzpicture}[thick,scale=0.6, every node/.style={scale=0.6},node distance=1.5cm]
    	\node (s1) [cell_r,label={[rotate=90,xshift=0.1cm,yshift=0.5cm]right:3*224*224}] {Conv\_K3};
	\node (s2) [cell_r_b, yshift=-1.5cm, xshift=-1.5cm, right of=s1,label={[rotate=90,xshift=0.1cm,yshift=0.5cm]right:16*112*112}] {MBE1\_K3};
        \node (a) [cell_r_g, e3, yshift=-1.5cm, xshift=-1.5cm, right of=s2,label={[rotate=90,xshift=-0.1cm,yshift=0.5cm]right:16*112*112}] {MBE3\_K5}; 
        \node (b) [cell_r_b, e3, yshift=-1.5cm, xshift=-1.5cm, right of=a,label={[rotate=90,xshift=-0.2cm,yshift=0.5cm]right:24*56*56}] {MBE3\_K3};
        \node (c) [cell_r_g, e3, yshift=-1.5cm, xshift=-1.5cm, right of=b,label={[rotate=90,xshift=-0.2cm,yshift=0.5cm]right:24*56*56}] {MBE3\_K5}; 
        \node (d) [cell_r_b,e3 , yshift=-1.5cm, xshift=-1.5cm, right of=c,label={[rotate=90,xshift=-0.2cm,yshift=0.5cm]right:40*28*28}] {MBE3\_K3};
        \node (e) [cell_r_g,e3 , yshift=-1.5cm, xshift=-1.5cm, right of=d,label={[rotate=90,xshift=-0.2cm,yshift=0.5cm]right:40*28*28}] {MBE3\_K5};
        \node (f) [cell_r_g, e3 , yshift=-1.5cm, xshift=-1.5cm, right of=e,label={[rotate=90,xshift=-0.2cm,yshift=0.5cm]right:40*28*28}] {MBE3\_K5};
        \node (g) [cell_r_g, e6, cell_se , yshift=-1.5cm, xshift=-1.5cm, right of=f,label={[rotate=90,xshift=-0.7cm,yshift=0.5cm]right:80*14*14}] {MBE6\_K5\_SE};
        \node (h) [cell_r_g, e3 , yshift=-1.5cm, xshift=-1.5cm, right of=g,label={[rotate=90,xshift=-0.2cm,yshift=0.5cm]right:80*14*14}] {MBE3\_K5};
        \node (i) [cell_r_g, e3 , yshift=-1.5cm, xshift=-1.5cm, right of=h,label={[rotate=90,xshift=-0.2cm,yshift=0.5cm]right:80*14*14}] {MBE3\_K5};
        \node (j) [cell_r_b, e6, yshift=-1.5cm, xshift=-1.5cm, right of=i,label={[rotate=90,xshift=-0.6cm,yshift=0.5cm]right:80*14*14}] {MBE6\_K3};
        \node (k) [cell_r_b, e6, cell_se , yshift=-1.5cm, xshift=-1.5cm, right of=j,label={[rotate=90,xshift=-0.6cm,yshift=0.5cm]right:112*14*14}] {MBE6\_K3\_SE};
        \node (l) [cell_r_b, e6 , cell_se, yshift=-1.5cm, xshift=-1.5cm, right of=k,label={[rotate=90,xshift=-0.6cm,yshift=0.5cm]right:112*14*14}] {MBE6\_K3\_SE};
        \node (m) [cell_r_b, e6 , cell_se, yshift=-1.5cm, xshift=-1.5cm, right of=l,label={[rotate=90,xshift=-0.7cm,yshift=0.5cm]right:160*7*7}] {MBE6\_K3\_SE}; 
        \node (n) [cell_r_b, e6 , cell_se, yshift=-1.5cm, xshift=-1.5cm, right of=m,label={[rotate=90,xshift=-0.7cm,yshift=0.5cm]right:160*7*7}] {MBE6\_K3\_SE};
        \node (bp) [cell_r,yshift=-1.5cm, xshift=-1.5cm, right of=n,label={[rotate=90,xshift=0cm,yshift=0.5cm]right:160*7*7}] {Conv\_K1};
        \node (pool_fc) [cell_r,yshift=-1.5cm, xshift=-1.5cm, right of=bp,label={[rotate=90,xshift=-1cm,yshift=0.5cm]right:960*7*7}] {Global Pooling + FC*2};
        \draw [arrow] (s1) -- (s2);
        \draw [arrow] (s2) -- (a);
        \draw [arrow] (a) -- (b);
        \draw [arrow] (b) -- (c);
        \draw [arrow] (c) -- (d);
        \draw [arrow] (d) -- (e);
        \draw [arrow] (e) -- (f);
        \draw [arrow] (f) -- (g);
        \draw [arrow] (g) -- (h);
        \draw [arrow] (h) -- (i);
        \draw [arrow] (i) -- (j);
        \draw [arrow] (j) -- (k);
        \draw [arrow] (k) -- (l);
        \draw [arrow] (l) -- (m);
        \draw [arrow] (m) -- (n);
        \draw [arrow] (n) -- (bp);
        \draw [arrow] (bp) -- (pool_fc);
        \draw [ultra thick, ccgray] (s1.south east)+(0.25,0.6) -- +(0.25,-2.2);
        \draw [ultra thick, ccgray] (a.south east)+(0.25,0.4) --+(0.25,-2.5);
        \draw [ultra thick, ccgray] (c.south east)+(0.25,0.4) --+(0.25,-2.5);
        \draw [ultra thick, ccgray] (f.south east)+(0.25,0.4) --+(0.25,-2.5);
        \draw [ultra thick, ccgray] (l.south east)+(0.25,-0.1) --+(0.25,-3);
        \draw [dashed, ccgray] (s2.south east)+(0.25,0.5) -- +(0.25,-2.3);
       \draw [dashed, ccgray] (n.south east)+(0.25,-0.1) -- +(0.25,-3.);  
    \end{tikzpicture}
}}
\caption{The Architectures of MoGA-A, B, C. Note E$x$\_K$y$\_SE means an expansion rate of $x$ for its expansion layer and a kernel size of $y$ for its depthwise convolution layer, SE for squeeze-and-excitation. Grey thick lines refer to downsampling points. Dashed lines separate the stem and end layers from the backbone.}
\label{fig:moga-architectures}
\end{center}
\end{figure*}

We are mostly comparable to the latest version of MnasNet \cite{tan2018mnasnet} and MobileNetV3 \cite{howard2019searching}, as we share the similar search space. Also, we use the same training and data processing tricks as in \cite{tan2018mnasnet} for complete training of stand-alone models. Note that with latency considered as one of the objectives, our generated models pay more attention to increase the number of parameters in order to gain higher performance, see detailed comparison results in Table~\ref{tab:comparison-imagenet}. We list all layers of MoGA-A in Table~\ref{tab:moga-a}, and illustrate the whole MoGA family in Figure~\ref{fig:moga-architectures}.

For a fair comparison, here we only consider single path models based on inverted bottleneck blocks. MoGA-A achieves a new state-of-the-art top-1 accuracy $75.9\%$, surpassing Proxyless-R Mobile (+$1.3\%$), MnasNet-A1 (+$0.7\%$), MnasNet-A2 (+$0.3\%$) with fewer FLOPs. MoGA-B obtains $75.5\%$, excelling MobileNetV3 at similar GPU speed. MoGA-C hits a higher accuracy with faster GPU speed, note it is slower on CPU, which otherwise will be treated as inferior by CPU-aware methods. Therefore, it's beneficial to fit models for specific hardware, indicating that even latency on other computing units and FLOPs are not ideal proxies. 

MoGA-A makes extensive use of large kernels  (4 layers with $7\times7$), which helps to enlarge receptive fields. Moreover, it mostly places large kernels on the stages with $14\times14$ input to reduce the latency cost. It also utilizes a large expansion rate after each downsampling stage to retain and to extract more useful features.

Interestingly for MoGA-B, the expansion rates across various layers mimic a sine curve. Like MoGA-A, it utilizes five $7\times7$ kernels to obtain a large receptive field. To cut down the latency cost, it places most of them in the $14\times14$ stage. Like FairNAS-A, it selects larger expansion rates right before downsampling operations.

Coincidentally, both MoGA-C and MobileNetV3-large simply contain $3\times3$ and $5\times5$ kernels only, even with same amount of such layers. While MobileNetV3-large prefers $5\times5$ operations in the tail of the model, MoGA-C chooses $3\times3$ instead. Besides, MoGA-C places $5\times5$ kernels in the middle and uses less squeeze-and-excitation operations. In such way, it better balances accuracy and latency cost.

\subsection{Mobile GPU Awareness Analysis}
We benchmark the inference cost for our three models both on mobile CPUs and GPUs. The result is shown in Figure~\ref{fig:pie-a-ops}. As for mobile GPUs, all models spend most of the time on 2D convolutions. MoGA-A and B spend the second most of the time on depthwise convolutions because they make extensive use of large kernels and expansion rates, whereas MoGA-C pays more attention to elementwise operations instead. Note all MoGA series invest more time on depthwise convolutions, contributing for faster speed and better performance.

It is worth noticing that how models exhibit a different behavior on mobile GPUs than on CPUs. For instance, vanilla convolutions and depthwise convolutions generally share bigger percentages on CPUs than on GPUs, while elementwise operations have a smaller percentage, as seen from Figure~\ref{fig:pie-a-ops}. Additionally, there is a discrepancy when running the same model with the different inference frameworks as well, which could call for a framework-aware solution, see Table~\ref{tab:comparison-imagenet}. Apart from the mobile framework we use, CPUs and GPUs differ on inherent microarchitectures, which puts hardware-specific requirements a must for the design of neural architectures.

\subsection{GPU Cost Analysis with More Mobile Devices}
\begin{figure}[ht]
\centering
{
\includegraphics[scale=0.6]{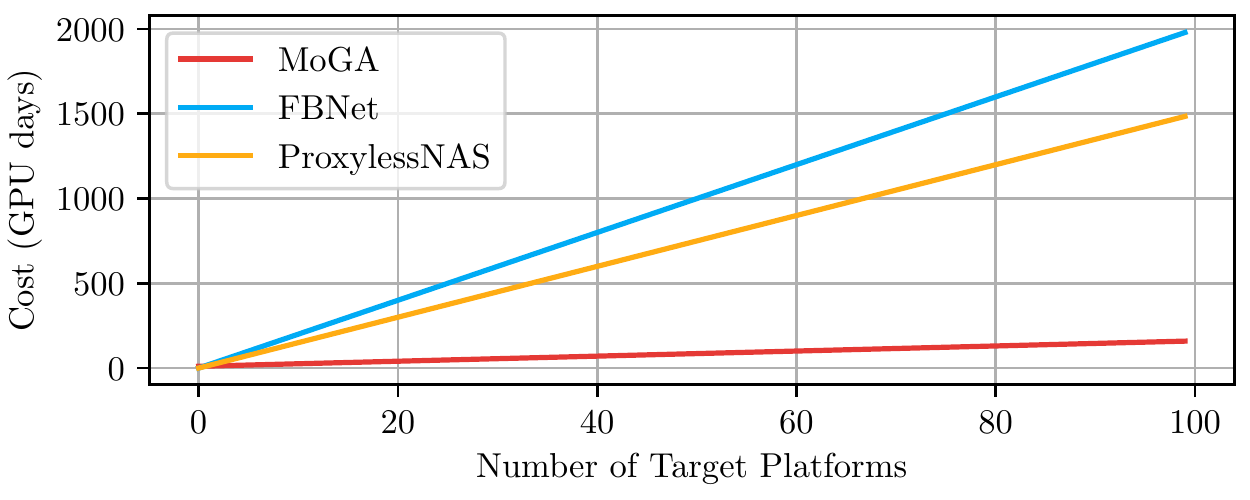}
}
\caption{Overall search cost vs. the number of target platforms.}
\label{fig:gpu-cost}
\end{figure} 
Given a target device, our overall search cost $c_{all}$ can be decomposed into two parts: $c_{super}$ for supernet training (10.5 GPU days) and $c_{search}$ for the NSGA-II pipeline. The latter estimates model accuracy by the supernet evaluator. Notably, there is no need to retrain the supernet when we design neural models for different mobile platforms. In contrast, the cost for most existing NAS methods, such as RL, EA and gradient descent, increases linearly with the number of platforms \cite{tan2018mnasnet,wu2018fbnet,cai2018proxylessnas,liu2018darts}. For $N$ platforms, our $c_{super}$ is amortized as $\frac{c_{super}}{N}$. When $N\ge22$, the overall cost $c_{all}$ reduces to less than 2 GPU days per platform. This benefit is better depicted in Figure~\ref{fig:gpu-cost}.

\begin{figure}[ht]
\subfigure{
\centering
{
\includegraphics[scale=0.6]{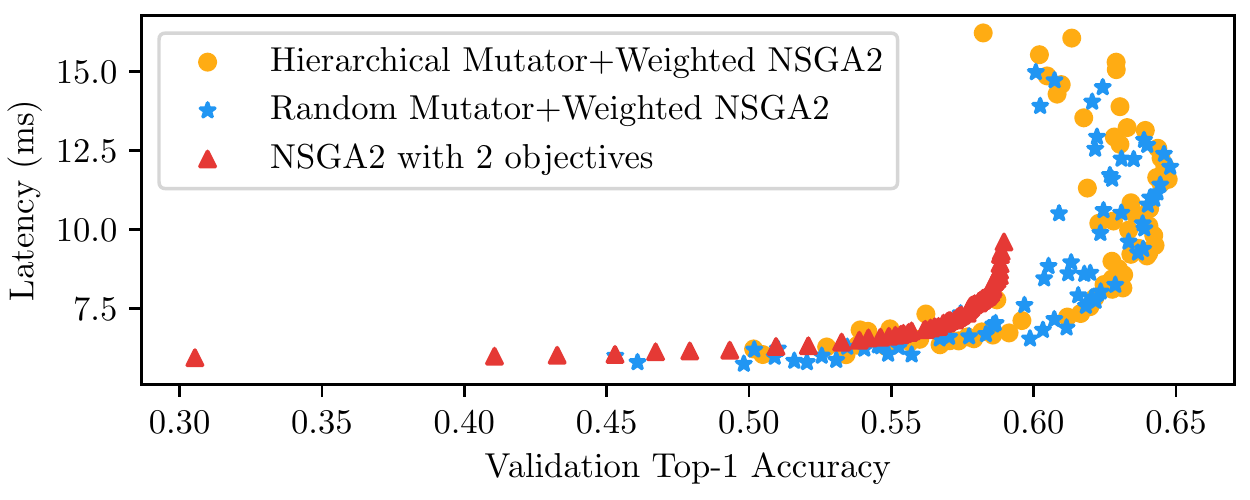}
}
}
\subfigure{
\centering
{
\includegraphics[scale=0.6]{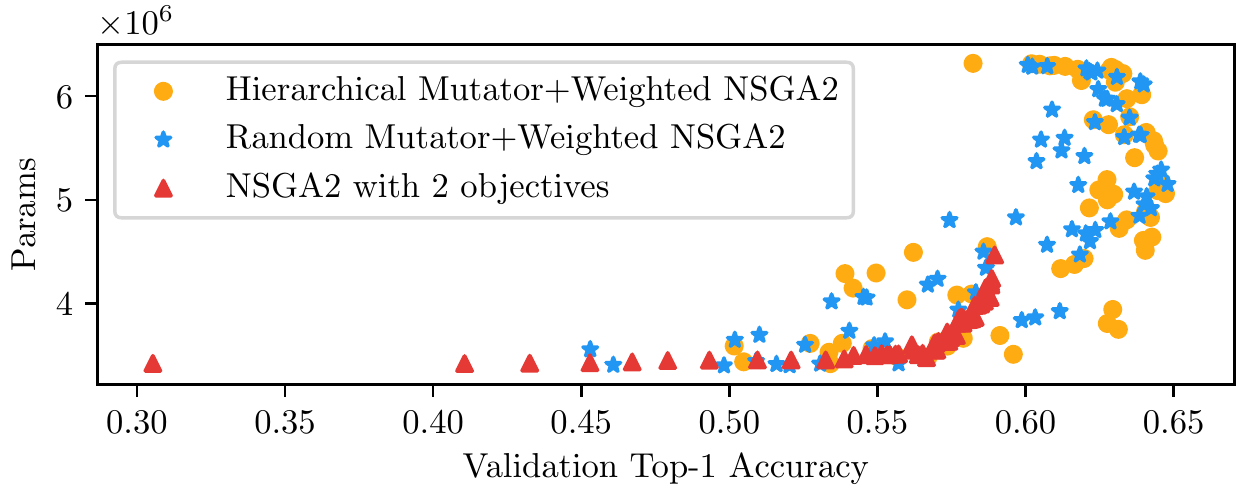}
}
}
\caption{Pareto Front of weighted NSGA-II with hierarchical mutator compared with that of a random mutator and of two objectives (accuracy, latency).}
\label{fig:ablation_best_models}
\end{figure}

\section{Results}
\label{sec:result}

\subsection{Ablation Study}
\subsubsection{Model Selection}

For the chosen weighted NSGA-II equipped with hierarchical mutation \cite{chu2019multi}, we compare it with the random mutation baseline, see in Figure~\ref{fig:ablation_best_models}. Plenty of models from the baseline are dominated by the hierarchical version, which attests that hierarchical mutation improves searching.

We compare the best elitists for Equation~\ref{eq: two-obj} and \ref{eq:three_objs}, which is shown in the upper part of Figure~\ref{fig:ablation_best_models}. The Pareto front formed by the two objectives is largely surrounded by those with three.  

While FairNAS states that a fair training can boost the rank relationship between the supernet predictor and stand-alone training, it also points out that it can be affected by initialization techniques and suboptimal training hyperparameters. For the latter, we empirically maximize the number of parameters as a compensation bonus. 

\subsubsection{Does it matter to use parameters as an objective?}

\begin{figure}[ht]
\centering
{
\includegraphics[scale=0.6]{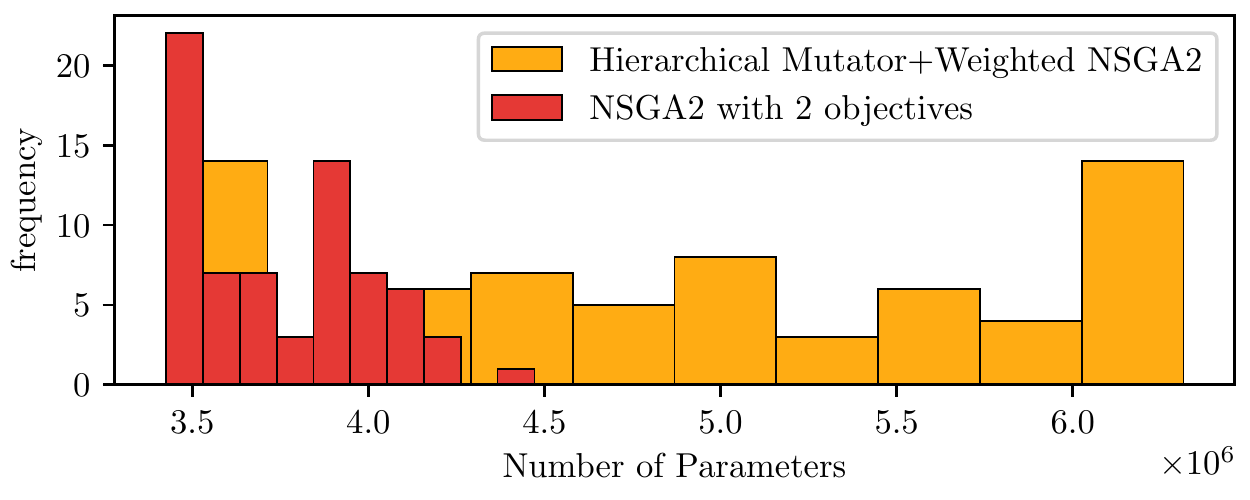}
}
\caption{Histogram on numbers of parameters of models from the last generation of weighted NSGA-II with hierarchical mutator,  compared with that of two objectives (accuracy, latency).}
\label{fig:histogram-params}
\end{figure}

%
Occam's Razor doesn't fit in this case, because in such mobile setting, a neural network is prone to underfitting instead of overfitting. If we consider minimizing the number of parameters, NSGA-II is then at the risk of excluding models with more parameters generation by generation. For evidence, we show the histograms of the number of parameters for the final elitists in Figure~\ref{fig:histogram-params}. 
 

\section{Conclusion}
\label{sec:conc}
To sum up, we have discussed several critical issues in mobile neural architecture design. \textbf{First}, we promote the first Mobile GPU-Aware (MoGA) solution, as in production, running networks on mobile GPUs are much preferred. \textbf{Second},  we adopt weighted fitness strategy to comfort more valuable objectives like accuracy and latency, other than the number of parameters. \textbf{Third}, our total search cost has been substantially reduced to 12 GPU days. Also, the trained supernet is once-for-all since the same supernet caters for all mobile contexts. It requires $o(1)$ search cost when applying to a new mobile device. \textbf{Last}, we employ an automated search approach in the search space adapted from MnasNet and MobileNetV3, which generates a new set of state-of-the-art architectures for mobile settings. In particular, MoGA-C hits 75.3\% top-1 ImageNet accuracy, which outperforms MobileNetV3 with competing mobile GPU latency at similar FLOPs and an equal number of parameters.

In the future, there will still be continuous interest to squeeze out better performance within limited hardware bounds, especially on targeted computing units. Also, balancing between architecture diversity and search space size will remain as a major topic, it also poses a challenge for searching algorithms when search space grows enormously.

\bibliographystyle{aaai}
\bibliography{aaai}

\end{document}